\documentclass[letterpaper, 10 pt, conference]{ieeeconf}
\IEEEoverridecommandlockouts
\usepackage[font=scriptsize]{caption}
\usepackage{url}
\usepackage[dvipsnames]{xcolor}
\usepackage{amsfonts}
\usepackage{amssymb,amsmath}
\usepackage{booktabs}
\usepackage{graphicx}
\usepackage{cases}
\usepackage{multirow}
\usepackage{tikz}
\usepackage{epstopdf}
\usepackage{algorithmic}
\usepackage{algorithm}
\usepackage[caption=false, skip=0pt, font=scriptsize]{subfig}
\usepackage{lipsum}
\usepackage{xargs}
\usepackage{mathtools,extarrows}
\makeatletter
\let\NAT@parse\undefined
\makeatother
\usepackage{hyperref}  
\hypersetup{
    colorlinks,
    linkcolor={green},
    citecolor={red},
    urlcolor={blue}
}

\newcommand{\Sajad}[1]{{#1}}
\newcommand{\Paul}[1]{{#1}}
\DeclareMathOperator*{\argmin}{\arg\!\min}

\newcommand{\vect}[1]{\mathbf{#1}}

\title{\Large \bf
BIT-VO: Visual Odometry at 300 FPS using Binary Features from the Focal Plane
}

\author{Riku Murai$^{1}$, Sajad Saeedi$^{2}$, Paul H. J. Kelly$^{1}$
\thanks{$^{1}$ Imperial College London, Department of Computing}%
\thanks{$^{2}$ Ryerson University}%
\vspace{-8 mm}
}

\begin{document}

\maketitle

\thispagestyle{empty}
\pagestyle{empty}


\begin{figure*}[htbp!]
\centering
\includegraphics[width=\linewidth]{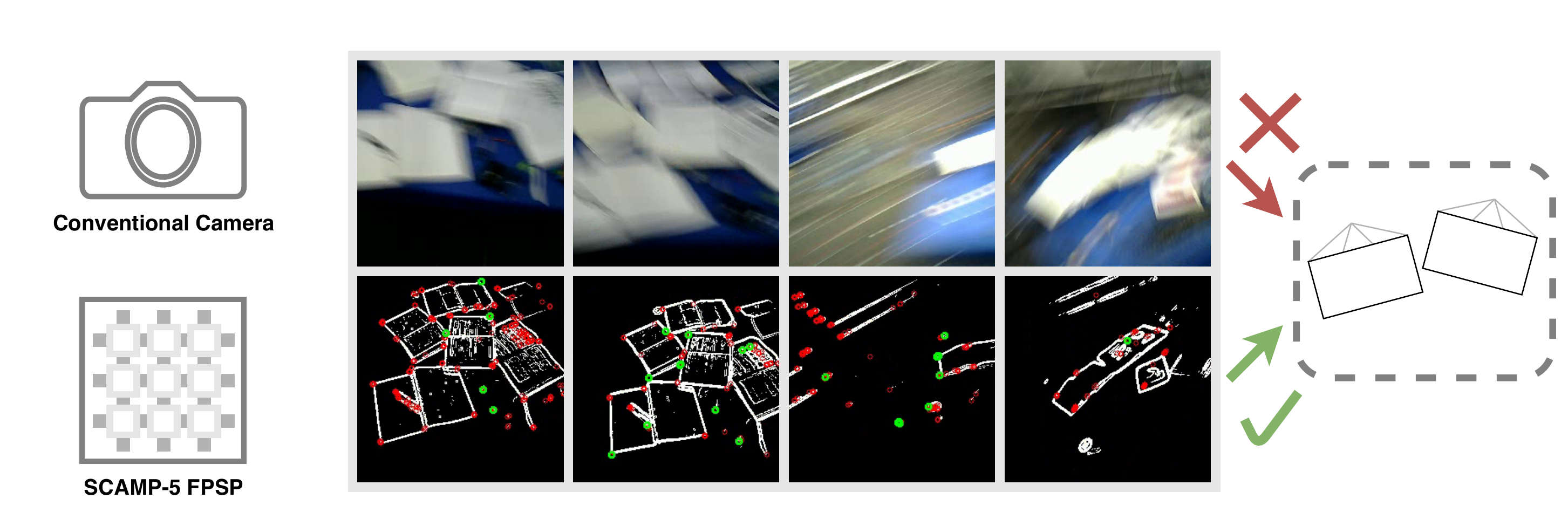}
\caption{Comparison of the data used by our proposed VO vs conventional VOs. Our system does not use intensity images (top row) but uses the binary edges and corners (bottom row) extracted by SCAMP-5 at 300 FPS. Notice that the edges, when extracted at a high frame-rate, are tolerant against motion blur, and are sharp even when the device is subject to violent motions. For the conventional camera (operating at 20 FPS)
, such motion severely blurs the images.}\label{fig:scamp-5-data}
\end{figure*}

\begin{abstract}
Focal-plane Sensor-processor (FPSP) is a next-generation camera technology which enables every pixel on the sensor chip to perform computation in parallel, on the focal plane where the light intensity is captured.
SCAMP-5 is a general-purpose FPSP used in this work and it carries out computations in the analog domain before analog to digital conversion.
By extracting features from the image on the focal plane, data which is digitized and transferred is reduced. As a consequence, SCAMP-5 offers a high frame rate while maintaining low energy consumption. 
Here, we present BIT-VO, which is, to the best of our knowledge, the first 6 Degrees of Freedom visual odometry algorithm which utilises the FPSP. Our entire system operates at 300 FPS in a natural scene, using binary edges and corner features detected by the SCAMP-5.
\end{abstract}

\section{INTRODUCTION}
Vision-based pose estimation algorithms, such as Visual Odometry (VO) and Visual Simultaneous Localization and Mapping (VSLAM), benefit from higher frame-rates; the two main benefits are a reduction in motion blur, 
and -- with smaller frame-to-frame motion -- a  faster optimisation convergence~\cite{handa2012real}.
However, state of the art algorithms operate at 30-80 frames per second (FPS), as increasing the frame-rate increases the volume of data to be processed. Even for fast VOs such as SVO~\cite{forster_svo:_2014, forster2016svo}, the author recommends a camera which operates at 40-80 FPS. Thus, to reduce the effect of motion blur caused by rapid camera movements, complex algorithms are required. DTAM~\cite{newcombe2011dtam}, which creates a dense 3D map, is one such example.  

This paper looks at the problem from a different perspective; if we can reduce the volume of data coming from the image sensor, we then have more time for pose estimation. Feature-based VO/VSLAM, such as PTAM~\cite{klein_parallel_2007} and ORB-SLAM~\cite{mur-artal_orb-slam:_2015} compute a sparse set of features from an image and operate using them. Unfortunately, feature extraction is often computationally expensive (ORB-SLAM requires 11ms for 1000 corners), and prevents increased frame-rate. The problem is that images are first transferred, and then the features are extracted. Instead, is there a way to stream just the relevant features from the image sensor?

Focal-plane Sensor-processor (FPSP)~\cite{zarandy_focal-plane_2011} is a general-purpose vision chip technology which allows user-defined computation in a highly parallel manner on the focal plane of the sensor at high frame rates.
\Sajad{For instance, SCAMP-5 can perform: High Dynamic Range Tone Mapping~\cite{martel2016real}, Depth from Focus~\cite{martel2017real}, and, FAST Keypoint Detection~\cite{chen2017feature} on the focal plane.} 
The low energy, high frame-rate nature of the FPSP, consuming only 1.23W even when operating at its maximum effective frame-rate of 100,000 FPS~\cite{carey_100000_2013}, makes the device appealing for high-speed operations.
The key to the efficiency of FPSPs -- in terms of both power consumption and frame-rate -- is the ability to reduce the amount of data transferred. As opposed to traditional camera sensors, FPSPs can perform image processing early in the pipeline to deliver a reduced volume of data to later stages -- \Paul{in this paper, just binarized corners and edges}. This reduces both bandwidth and energy consumption. 


Similar to FPSPs, event-based cameras are another low power, low latency camera technology, which output an asynchronous stream of intensity changes~\cite{Lichtsteiner2008}.
Many VO/VSLAM algorithms have been implemented using event cameras~\cite{gallego2019eventbased}, however, the bandwidth of data transferred is proportional to the manoeuvre speed -- fast motion requires more processing. On the other hand, an FPSP can be programmed to output data at a consistent data rate, thus there is no significant fluctuation in the amount of data transferred under any sort of motion.

The objective of this work \Sajad{is} to investigate this approach in estimating the pose of the FPSP in 3D space, predicting motions with all 6-Degrees of Freedom (DoF).
The contributions of our work are: 
\begin{itemize}
    \item An efficient BInary feaTure Visual Odometry, BIT-VO which operates at over 300 FPS. Using no intensity information, our proposed method is able to accurately track the pose, even under difficult situations where the state of the art monocular SLAM fails.
    \item A robust feature matching scheme, which uses our novel binary-edge based descriptor. Using a small, 44-bit descriptor, our system is able to \Paul{track the noisy feature data computed on the focal plane in the SCAMP-5 FPSP image sensor itself.} 
    \item Extensive evaluation of our system against measurements from a motion capture system, including difficult scenarios such as violently shaking the device 4-5 times in a second.
\end{itemize}

The remainder of the paper is organised as follows.
Section~\ref{sec:background} describes the SCAMP-5 FPSP and reviews related work.
Section~\ref{sec:overview} provides an overview of our system, together with the notations used. 
Section~\ref{sec:feature} and Section~\ref{sec:vo} explain the proposed visual odometry algorithm.
Section~\ref{sec:experiments} details our experimental results.
Finally, Section~\ref{sec:conclusion} concludes our work and discuss directions for the future.

\section{BACKGROUND}\label{sec:background}
\Sajad{This section provides a background and literature review on two topics: the SCAMP-5 FPSP and VO using unconventional vision sensors.}
\subsection{\Sajad{SCAMP-5 FPSP}}
An FPSP is a general-purpose vision chip, where sensors and processor are integrated together on the same silicon~\cite{zarandy_focal-plane_2011}.
Our work uses the SCAMP-5 vision system~\cite{dudek2005general}, which is an FPSP with a resolution of $256\times256$ pixels.
On the focal plane, each one of the 65,536 pixels \Paul{combines a photodiode with a Processing Element (PE).}
The device is programmable in a Single Instruction, Multiple Data (SIMD) fashion, where all of the PEs execute the same instruction.
Each one of the PEs is capable of storing local data using 7 analog and 13 1-bit registers. Each PE can also perform simple computations such as logical and arithmetic operations.
The arithmetic operations occur in the analog domain and directly on the analog registers without the need for digitization~\cite{carey_100000_2013}.
The nature of analog computation results both in arithmetic operation incurring noise, and values stored on analog registers degrading over time~\cite{carey_mixed_2013}.
A PE can communicate with neighbours to its north, east, west or south by copying its register value.
Once the computation is complete, data can be read out from the device in the form of coordinates, binary frames, analog frames, or global data (e.g. regional summation)~\cite{dudek2005general}. In particular, coordinates can be read out using event-readout, \Paul{where the cost, in time and energy,} is proportional to the number of events rather than the image dimension.
This flexibility allows data reduction to occur on the focal plane~\cite{carey_100000_2013}, and, together with highly parallel instruction execution, FPSPs are able to perceive and process visual information at a very high frame-rate.

When developing algorithms for the SCAMP-5 vision system, not only is there \Paul{inaccuracy and noise} in the analog computation, but there are resource constraints which make the porting of a computer vision algorithm rather complicated.
\Paul{The instruction set is limited; a wide range of logical operators are available for the 1-bit registers but only simple arithmetic operators (e.g. addition and subtraction) are available for the analog registers. However, there are few available registers. Furthermore,
there is no global memory.} A PE's only means of communication is to share data with its adjacent neighbours.
Although these factors provide challenges, there have been successful implementations of complex algorithms, for example, a convolutional neural network capable of classifying handwritten digits at 2250 FPS~\cite{wong_analog_2018}, ternary
weight CNNs~\cite{bose2019camera}, and a tracker for a ground target from a UAV at over 1000 FPS~\cite{greatwood2017tracking}.


\subsection{Visual Odometry Using Unconventional Vision Sensors}
\Sajad{The term visual odometry (VO) was first coined by Nister et al. \cite{Nister2011VO}. VO is the process of determining the egomotion of a sensor using visual information. Many algorithms have been proposed for VO using conventional vision cameras, for a review, see \cite{Scaramuzza2011RAM}.}
With the introduction of 
unconventional visual sensing technologies such as FPSPs and event cameras, the potential use of such sensors in VO is an active field of research.

\Sajad{Few works exist performing VO using FPSPs and none of them are 6-DoF.} 
A 4-DoF VO algorithm using a direct method was proposed in~\cite{debrunner2019Multiprog}. This approach divides the image into $N$ tiles and estimates an optic flow of each of the tiles efficiently on the focal plane.
These vectors are decomposed using ordinary least squares to predict the yaw, pitch, roll and z-axis motion of the device. The computation all occurs on the SCAMP-5, allowing the algorithm to operate at 400-500 FPS.
Another 4-DoF VO algorithm which instead uses a feature-based approach was proposed in~\cite{bose_visual_2017}. The edge features are extracted from the captured image and are aligned against a keyframe using image shifting, scaling, and rotation. All of the image manipulations occur on the focal plane. By measuring the amount of shift, scaling and rotation required to align two images, an estimate of the yaw, pitch, roll and motion in the z-axis is obtained.
Given sufficient lighting, the algorithm is capable of operating at over 1000 FPS.
Both methods achieve high frame-rate by performing all of the computation on the SCAMP-5 device; however, they are limited to 4-DoF tracking, restricting their use cases to platforms which are mechanically constrained to motion in one direction.

\Sajad{Many algorithms along with benchmarking datasets have been proposed for VO using event cameras. For further information about algorithms, challenges and future directions, see the recent review \cite{gallego2019eventbased} by Gallego et al. 
Event cameras have also been combined with frame-based cameras \cite{Brandli2014DAVIS} to improve VO and other algorithms by detecting features such as edges and corners \cite{gallego2019eventbased}. While the combined approach shares similarities with the FPSPs architecture, a major difference is that FPSPs is capable of processing data in-situ.}  




\section{System Overview}\label{sec:overview}
\begin{figure}
    \centering
    \includegraphics[width=\linewidth]{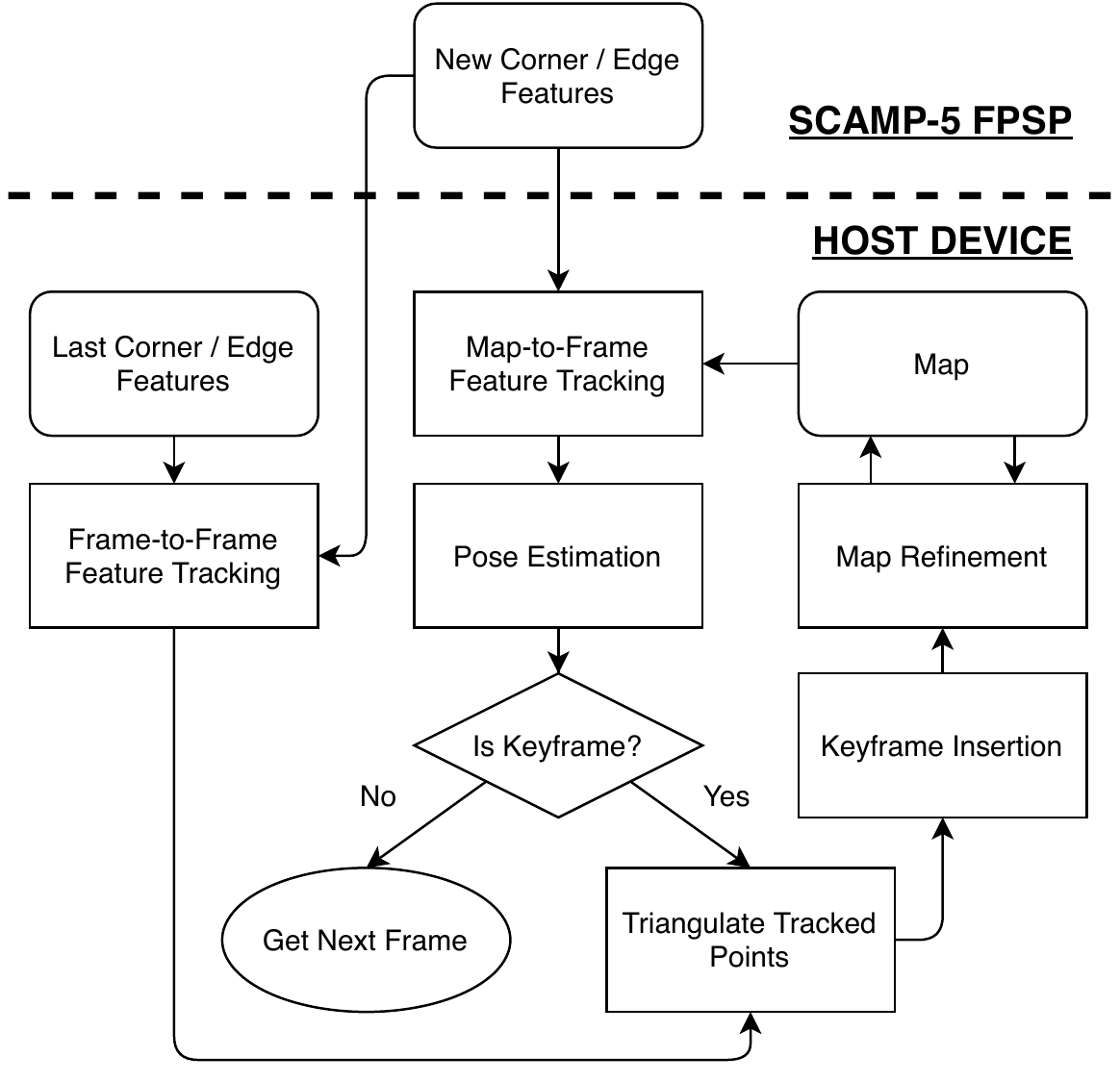}
    \caption{Tracking and Mapping pipeline.  \Sajad{The pipeline runs on an FPSP and a host device, minimising data flow from the sensor to host device,}}
    \label{fig:visual-odometry-pipeline}
\end{figure}
Our main contribution is a 6-DoF monocular visual odometry which operates in real-time at 300 FPS.
An overview of our system flow is summarised in Fig.~\ref{fig:visual-odometry-pipeline}. The initialisation is omitted for simplicity. Feature extractions are performed on the SCAMP-5, while feature tracking and VO operates on the host device which is, for example, a consumer grade laptop.
The system operates only on the binary edge image and corner coordinates, thus no pixel intensity information is ever transferred (Fig.~\ref{fig:scamp-5-data}). Only using the limited information, we demonstrate that it is possible to create a VO system which is robust against rapid motion.

\subsection{Notations}
The world frame is represented with $\vect{w}$ and the camera frame is represented with $\vect{c}$.
The position $\vect{p}$ in world frame is denoted as $\vect{{}_{w}p}$.
The rigid-body transformation $\vect{T_{c,w}}\in\vect{SE(3)}$ expresses the transformation from the world to the camera frame.
This allows a point in the world frame $\vect{{}_{w}p}$ to be mapped to the camera frame by $\vect{{}_{c}p} = \vect{T_{c,w}}\,\vect{{}_{w}p}$.

\section{Feature Detection and Matching}\label{sec:feature}
This section outlines how features is detected on the FPSP device, and how these features are matched against previous ones on the host device.
\subsection{Feature Detection}
Corner and edge features are computed on an FPSP, and it operates at a high frame-rate of 330 FPS.
FAST Keypoint Detector~\cite{rosten2006machine} is used for the corner detection. For the edge detection, the magnitude of the image gradient is thresholded to find edges~\cite{bose_visual_2017}. 
An existing implementation of FAST Keypoint Detector for the SCAMP-5~\cite{chen2017feature} is used in our work, although suppression of features is disabled as removal of the features is not repeatable. 
Unfortunately, performing repeatable suppression techniques such as non-maximal suppression is difficult due to the noisy analog computation on SCAMP-5. The noisy computation leads to not only incorrect inequality comparisons but also to incorrect computation of the compared values.
For every incoming frame, SCAMP-5 detects at most 1000 corner features which are read-out using an event-readout. 
For the binary edge image, the whole $256\times256$ bit image is transferred, rather than coordinates.  
In SCAMP-5, coordinates are expressed as an 8-bit pair, hence, event-readouts are only efficient if the number of events $N_{events} < 4096$. This is only $6.25\%$ of all the available pixels, and, we have found that in a typical indoor scene, around 10-15\% of the pixels are classified as an edge feature.

\subsection{Feature Matching}
\begin{figure}[!t]
    \centering
    \includegraphics[width=\linewidth]{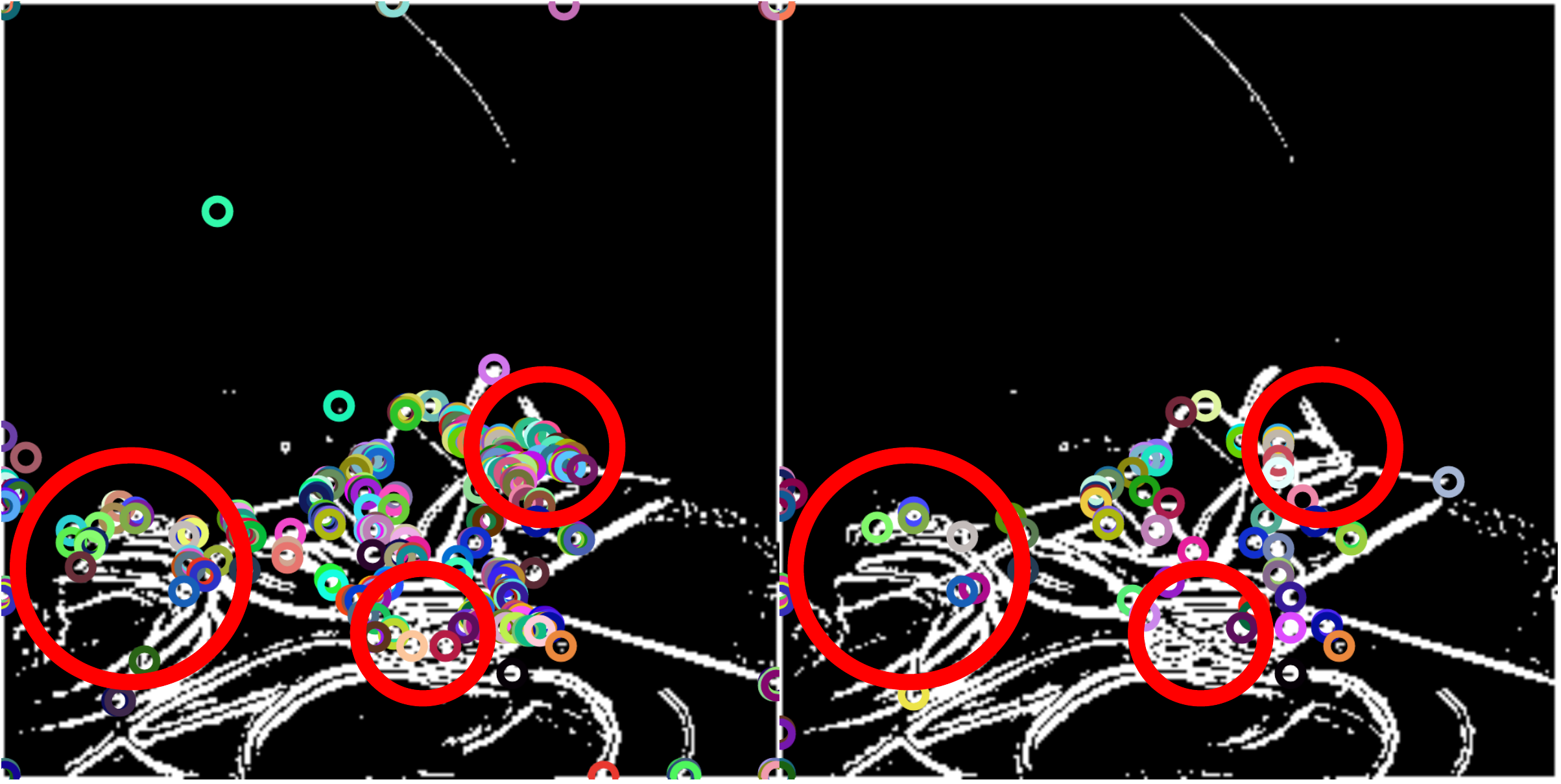}
    \caption{Illustration of the effect of noisy analog computation. Between two consecutive frames, many corners appear and disappear. The device was mounted on a tripod to ensure stability of the device across multiple frames.}
    \label{fig:consecutive_frames}
\end{figure}

The task of matching the corner features are challenging for two reasons: a) Feature extraction suffers from noise in analog computation, b) multiple features are extracted per visual corner.
Since SCAMP-5 performs computations using an analog circuit, corners are not reliably extracted in every frame (Fig.~\ref{fig:consecutive_frames}), causing incorrect association if one uses a naive method such as nearest neighbour matching.
For each incoming frame of features, one may search a small local neighbourhood to avoid establishing correspondence if a corner is not extracted because of noise; however, due to (b), it will misidentify correspondence with an incorrect corner feature. This will build up error and result in many incorrect associations, and thus poor visual odometry.

\subsubsection{Local Binary Descriptors from Edges}
\begin{figure}
    \centering
    \includegraphics[width=0.55\linewidth]{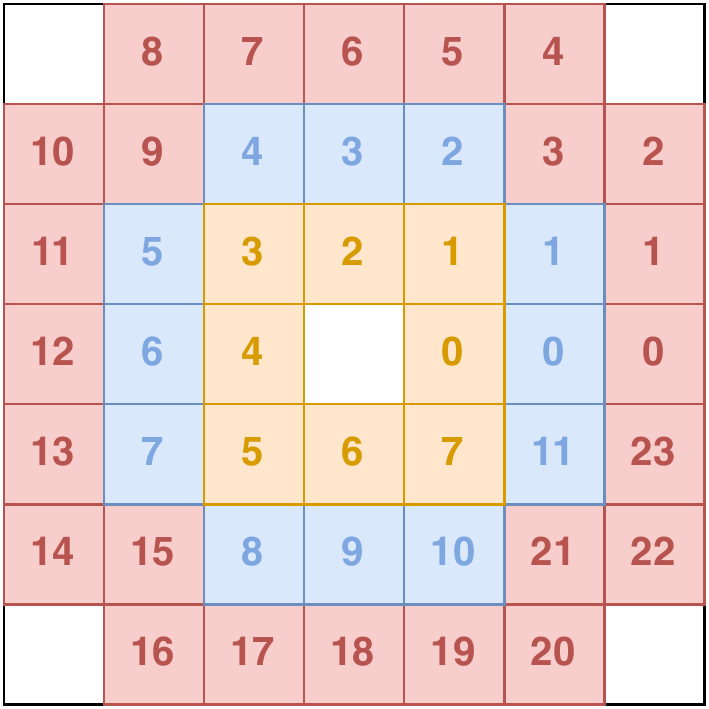}
    \caption{Descriptor sampling pattern. Different colours denote a different ring, and indices correspond to the bit index.}
    \label{fig:feature-pattern}
\end{figure}

To establish robust correspondences across multiple feature frames, we propose a feature descriptor which only uses the local binary edge information. 
Our descriptor is tiny -- only 44-bit in length thus is space-efficient and is fast to compute. Unlike other binary descriptors~\cite{ojala2002multiresolution, calonder2010brief, leutenegger2011brisk}, we do not have access to the image intensity information.
As shown in Fig.~\ref{fig:feature-pattern}, three independent rings $\{r1, r2, r3\}$ are formed around the corner of interest, and each of the rings element stores one bit of binary edge image below. A $7\times7$ patch is used as it fits in a single 64-bit unsigned integer. This allows the patch data to be converted into rings efficiently using bitwise manipulation only. 
To add rotation invariance to the descriptor, the orientation of each of the features is computed. Assuming a coordinate frame with the origin set to a corner feature of interest, the intensity gradient magnitude $G(x, y)$~\cite{rosin1999measuring} is used to compute the orientation  
\begin{equation}\label{eq:grad}
    \theta = \tan^{-1}\frac{\sum_{x,y} yG(x, y)}{\sum_{x, y} xG(x, y)}
\end{equation}
where and $x, y$ are the coordinates of the $7\times7$ patch.
Since the gradient image is binarized, Eq.~\ref{eq:grad} is approximated by
\begin{equation}
    \theta = \tan^{-1}\frac{\sum_{x,y} yB(x, y)}{\sum_{x, y} xB(x, y)},
\end{equation}

where $B(x, y)$ is 1 if image point $(x, y)$ is classified as an edge, and 0 otherwise.
The rotation invariance is achieved by bit-rotations of the rings independently~\cite{ojala2002multiresolution}, based on the orientation $\theta$.
At each ring, the number of bits to rotate is determined by $ 
    \textit{rotate\_by}(\theta, r) = \left\lfloor \theta \cdot \#r/360 \right\rfloor
$
where $r \in \{r1, r2, r3\}$ and $\#r$ is length of the ring.
Finally, the descriptor is computed by taking the disjunction of $\{r1, r2, r3\}$ after shifting $r1$ by $\#r2 + \#r3$ and $r2$ by $\#r3$.
The descriptors are compared against each other using the Hamming distance, which is performed efficiently using SSE instructions. Although our descriptors are not scale-invariant, they are sufficient for small indoor environments. 

\subsubsection{Frame-to-Frame Matching}
The high frame-rate of our system enables efficient frame-to-frame feature matching.
Given frames, $\{F_1, \ldots, F_n\}$, a local neighbourhood around a feature in $F_i$ is matched  against features in $F_{i+1}$. Similarly, features in $F_{i+1}$ are matched against features in $F_{i+2}$. By following these matches, features in $F_i$ can be matched against any arbitrary frames, assuming that they stayed in sight.
This enables feature tracking to take advantage of the small inter-frame motion.
By searching a small radius of $3-5$ pixels, a feature which minimises the Hamming distance is selected as a candidate.
If the descriptor distance to the candidate exceeds a threshold, the candidate does not form a match. In our implementation, threshold is set to 10.
\subsubsection{Map-to-Frame Matching}
All of the visible map points are back-projected onto the image plane to find correspondences. Again, only a small radius is searched.
Map points are observed by multiple keyframes, thus, they store multiple descriptors each. 
Similar to ORB-SLAM~\cite{mur-artal_orb-slam:_2015}, the most descriptive descriptor is selected by finding a descriptor which minimises the median distance to all others.


\section{VISUAL ODOMETRY}\label{sec:vo}

Our VO system uses information obtained through feature tracking to predict the 3D point-cloud structure of the scene and the pose of the SCAMP-5.
Like PTAM~\cite{klein_parallel_2007}, localisation and mapping are interleaved.
The 3D map points are generated through triangulation of features, and the pose of the SCAMP-5 is estimated through minimisation of the reprojection error. The non-linear optimisation is solved using the Levenberg-Marquardt algorithm, implemented using the Ceres Solver~\cite{agarwal_ceres_nodate}.
Since the inter-frame motion is small, the non-linear optimisation is fast to converge, requiring at most 10 iterations.

\subsection{Pose Estimation}
Given a set of 3D map points and its correspondences on an image plane, poses can be estimated by minimising the reprojection error, which can be formulated as~\cite{strasdat_visual_2012}:
\begin{equation}\label{eq:motion_only_ba}
  \vect{T_{c,w}} = \argmin_{\vect{T_{c,w}}}\frac{1}{2} \sum_{i} \rho \left(\| \vect{u}_i - \pi(\vect{T_{c,w}}\,\vect{{}_{w}p}_i)\|^2\right)
\end{equation}
Where the error between the projected 3D points $\pi(\vect{T_{c,w}}\,\vect{{}_{w}p}_i)$ and the corresponding feature coordinates $\vect{u}_{i}$ are minimised.
$\rho(\cdot)$ is the Huber loss function which reduces the effect of outlying data~\cite{zhang_parameter_1997}.
Unlike PTAM~\cite{klein_parallel_2007} or ORB-SLAM~\cite{mur-artal_orb-slam:_2015}, the velocity model is not used in pose estimation.
At such high frame-rate, inter-frame motion is small, thus the previous pose is a sufficiently good estimate of the current position. Furthermore, the addition of velocity model leads to worse initialisation if the camera motion violates this assumption, which occurs often during violent motion.

\subsection{Map Refinement}
Every map point keeps a reference of the keyframes that it was observed by.
These relationships form a graph, which is used in the structures-only bundle adjustment, where the pose estimates for each of the keyframes remain fixed, and only the positions of the map points are optimised. This is solved robustly using the Huber loss and at the end of the optimisation, map points are removed if their residual exceeds the Huber functions tuning constant.

\subsection{Initialisation}
The 5-point algorithm~\cite{nister_efficient_2004} with RANSAC~\cite{fischler_random_1981} is used to perform bootstrapping. This gives a relative pose estimate, which is used to triangulate the initial 3D map. Features in the reference frame are tracked using frame-to-frame tracking until there are sufficient disparities. Disparities are computed by taking the median of the features pixel displacements. If it is greater than 20 pixels, relative pose estimation and triangulation is attempted. Upon triangulation, if any 3D map point has a parallax of less than 5 degrees, or is behind of either of the two cameras, they are removed from the map.
Once over 100 map points are successfully triangulated, the system is initialised.

\subsection{Keyframe Selection}
To select which frames are suitable as a keyframe, similar to PTAM~\cite{klein_parallel_2007} and SVO~\cite{forster_svo:_2014, forster2016svo}, the selection process is based on the displacement of the camera relative to the depth of the scene. A keyframe is inserted when all of the following conditions are satisfied: 
a) At least 200 frames have passed since the previous keyframe insertion, 
b) at least 50 features are tracked, and 
c) Euclidean distances between the current frame and all the other keyframes are greater than 12\% of the median scene depth. 
When a frame is selected as a keyframe, first, 2D-3D correspondences are established through the back-projection of the map points into the image plane. This links the map point to the keyframes that observed it. 
For the features which are not yet triangulated, Frame-to-Frame tracker is inspected to see if there are any successful matches which satisfy the epipolar constraint.
If not many matches are found ($< 30$ matches), brute-force matching of the features is performed between the current and the last keyframe. This process ensures that a sufficient number of map points are created at every keyframe insertion.


\begin{figure}[!t]
    \centering
    \includegraphics[width=\linewidth]{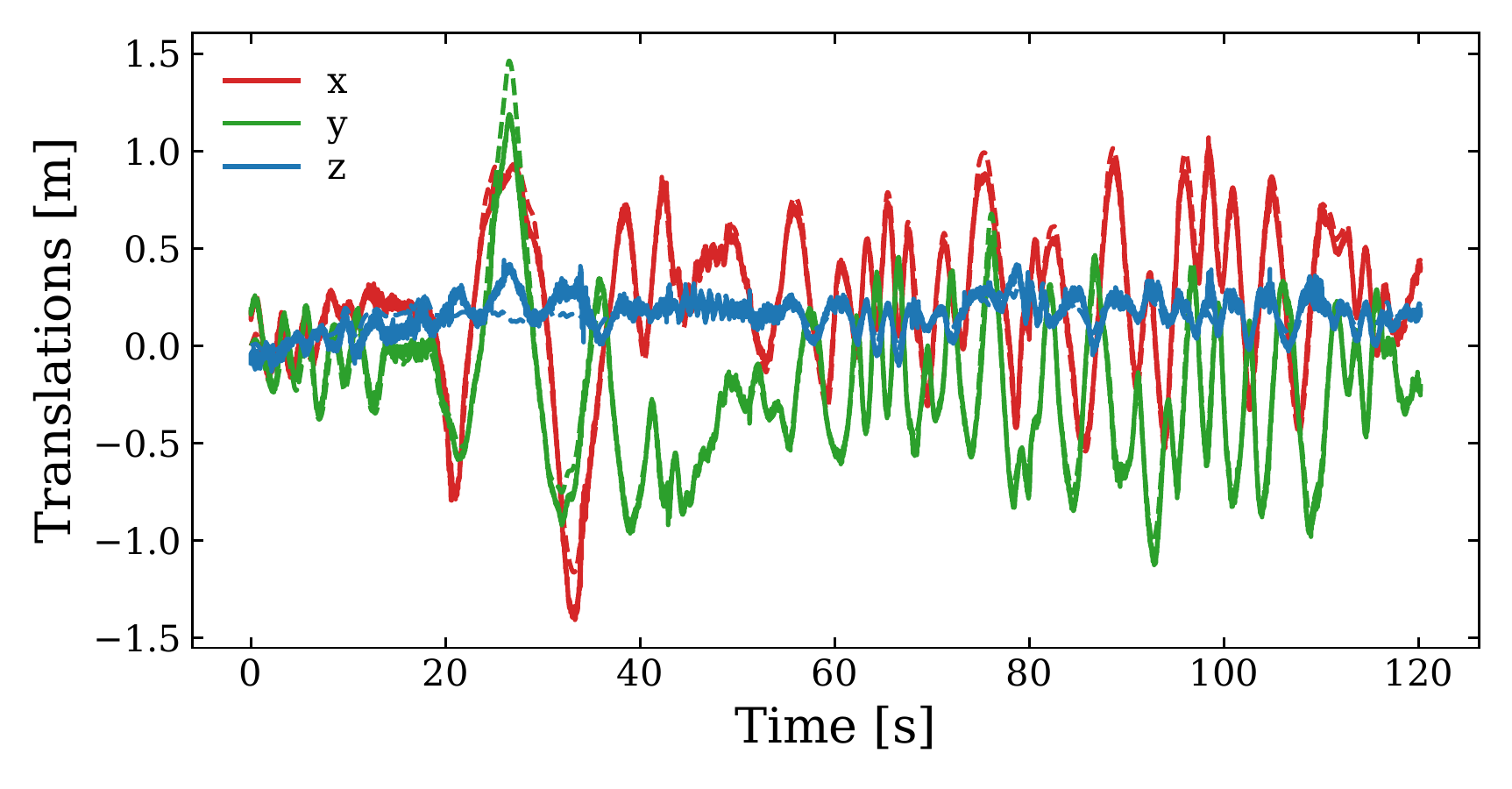}
    \caption{Estimated x, y, z translations for ``Long" sequence. Solid lines show our estimate and dotted lines are the ground truth.}
    \label{fig:long-translation}
    \centering
    \includegraphics[width=\linewidth]{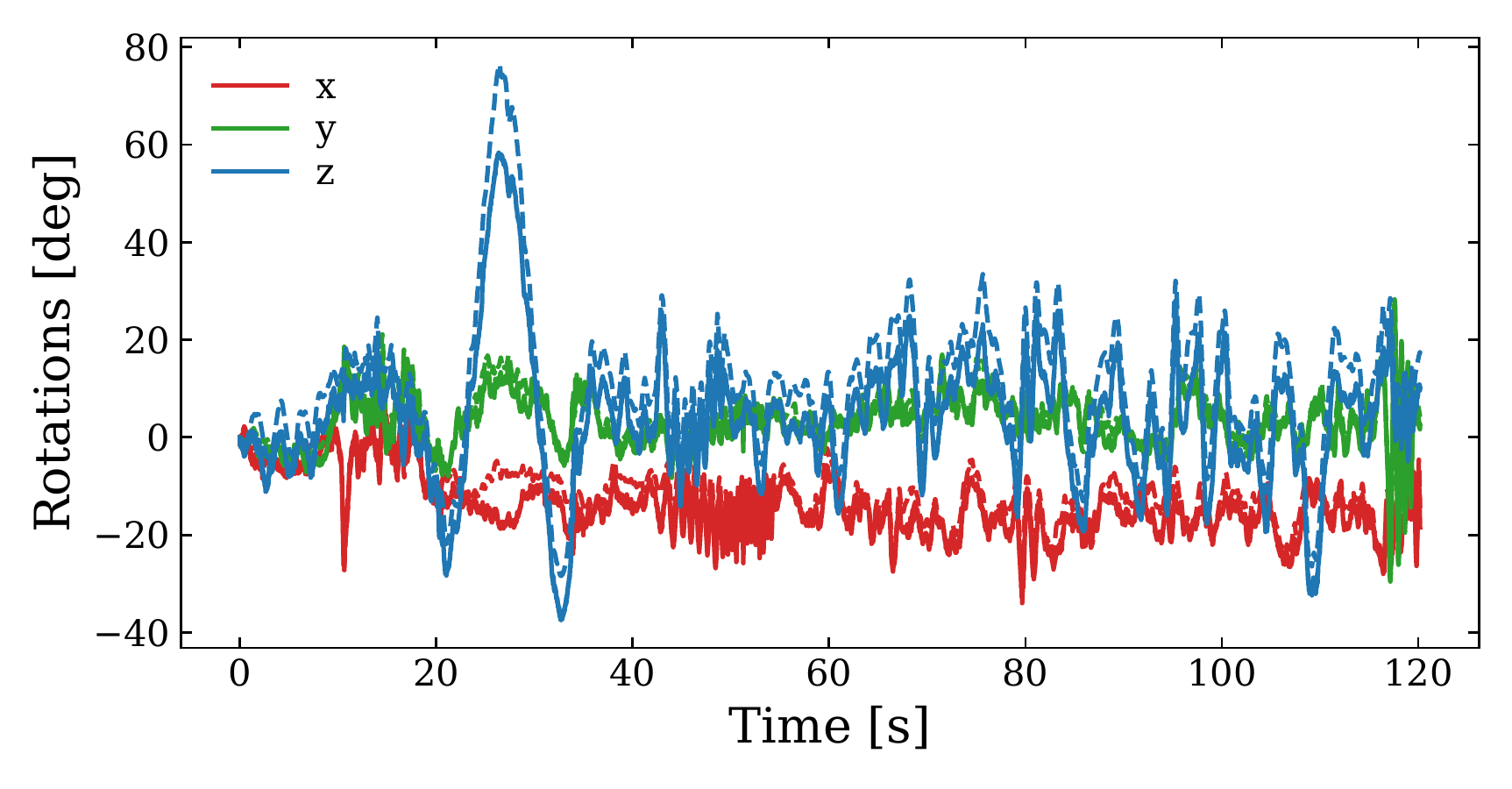}
    \caption{Estimated x, y, z rotations for ``Long" sequence. Solid lines show our estimate and dotted line are the ground truth.}
    \label{fig:long-orientation}
\end{figure}
\begin{figure}[!th]
    \centering
    \includegraphics[width=\linewidth]{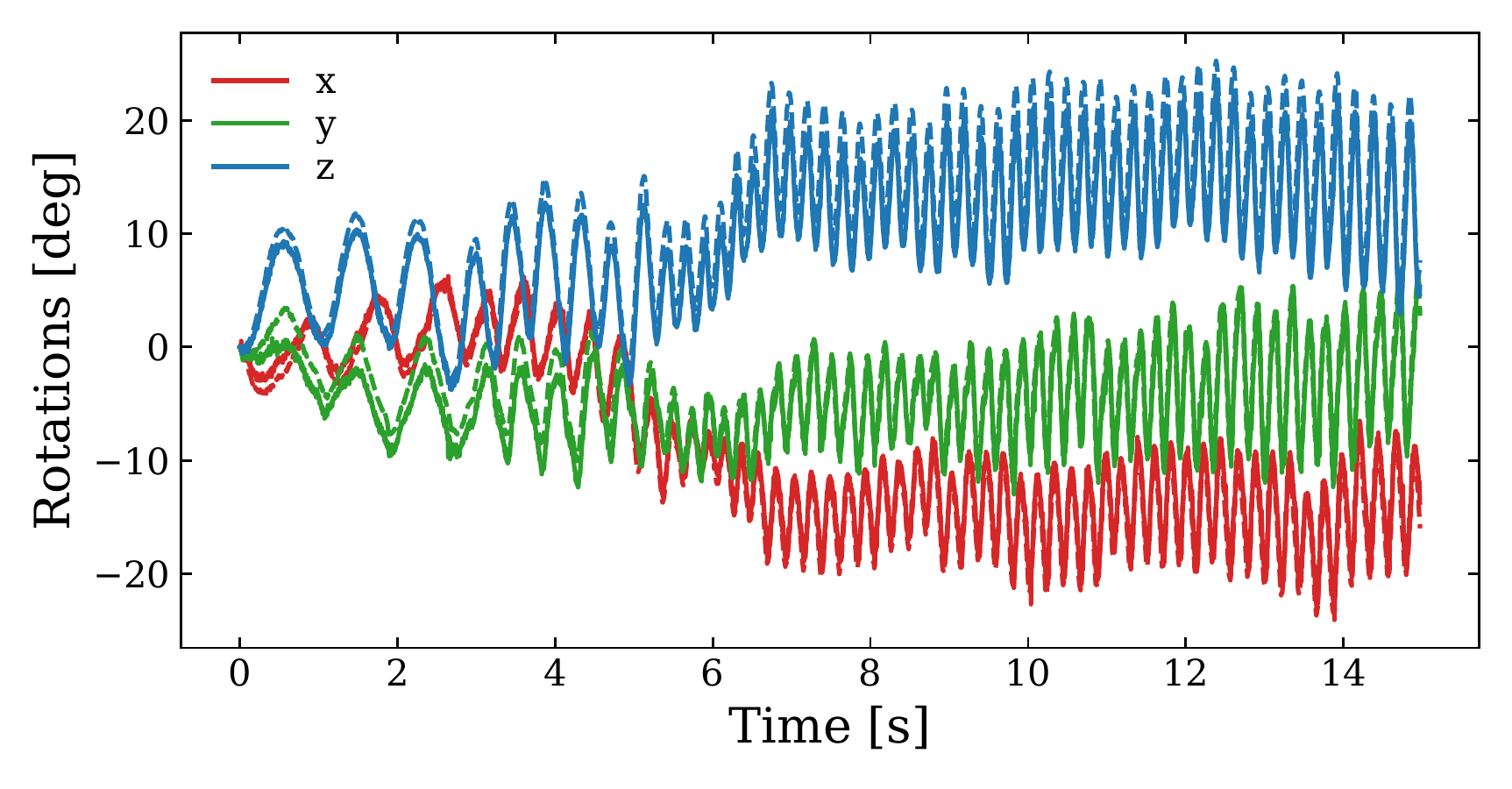}
    \caption{Estimated x, y, z rotations for ``Rapid Shake" sequence. Solid lines show our estimate and dotted line are the ground truth.}
    \label{fig:fast-orientation}
    \centering
    \includegraphics[width=\linewidth]{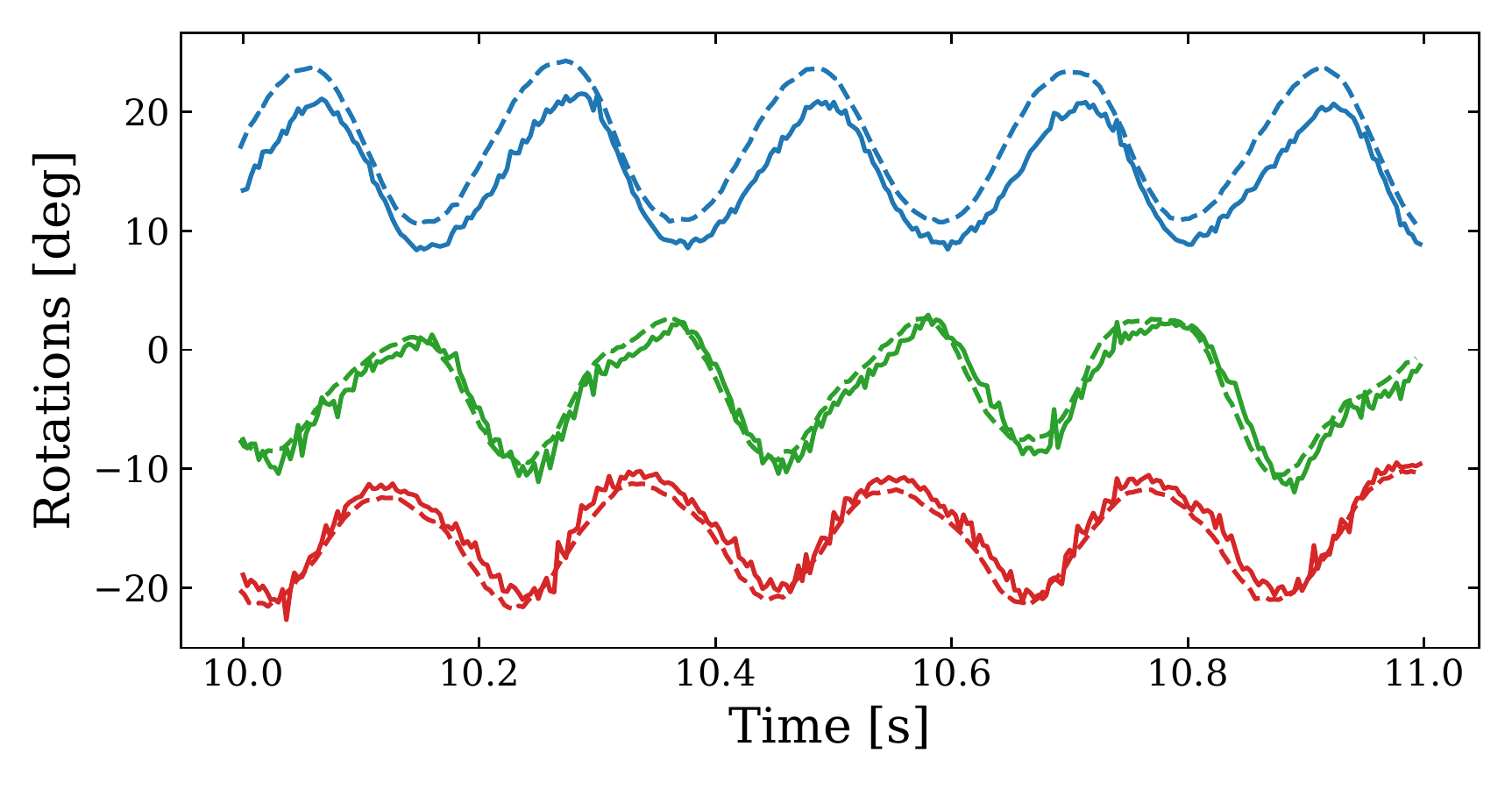}
    \caption{Close-up view of rotation estimates for ``Rapid Shake" sequence. Our proposed method is capable of tracking rapid rotation accurately.}
    \label{fig:fast-orientation-zoom}
\end{figure}
\begin{figure}[!ht]
    \centering
    \includegraphics[width=\linewidth]{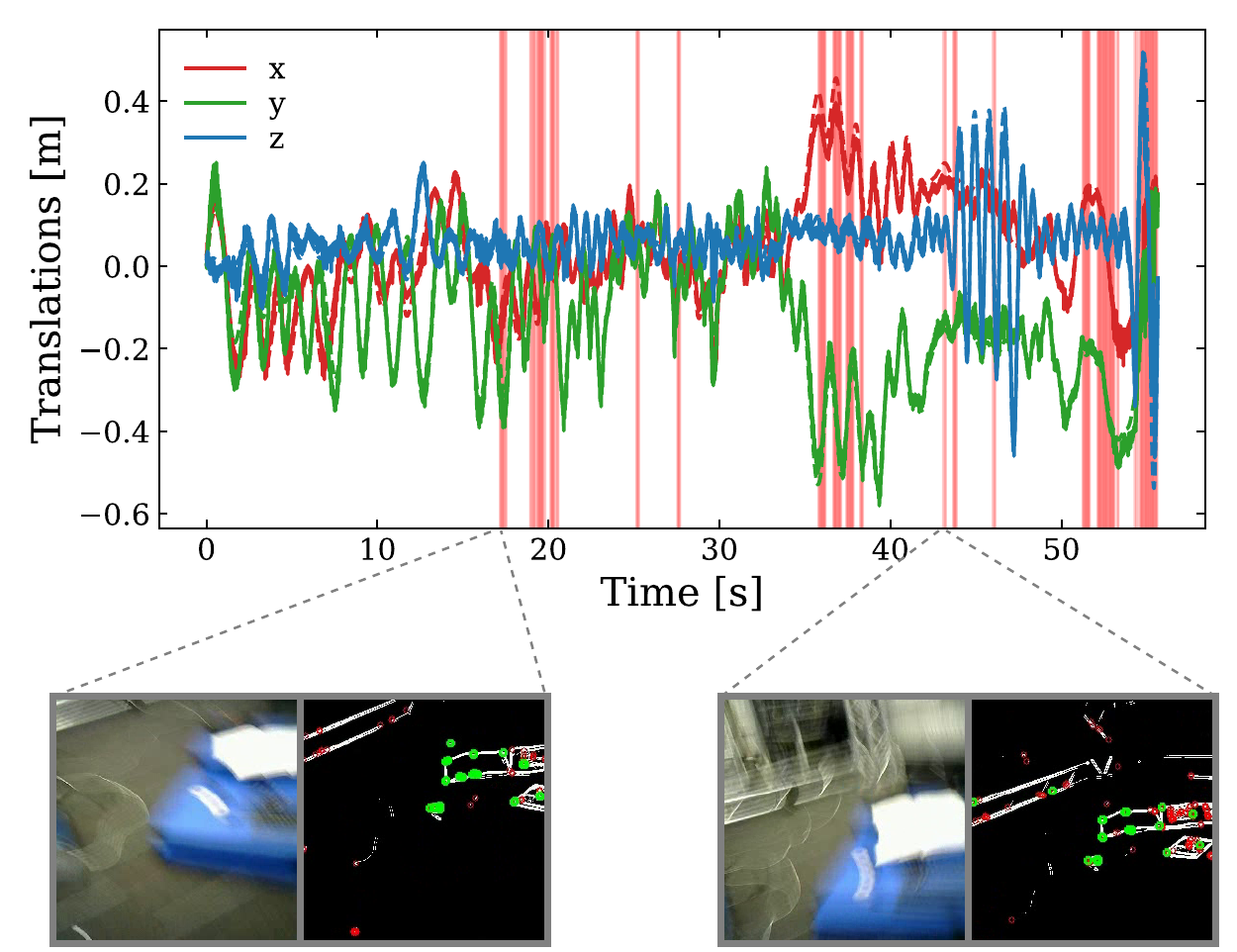}
    \caption{Estimated x, y, z rotations for ``Jumping" sequence. Solid lines show our estimate and dotted line are the ground truth. The pink region indicates that the ORB-SLAM lost track due to rapid motion.}
    \label{fig:orb-fail-translation}
    \centering
    \includegraphics[width=\linewidth]{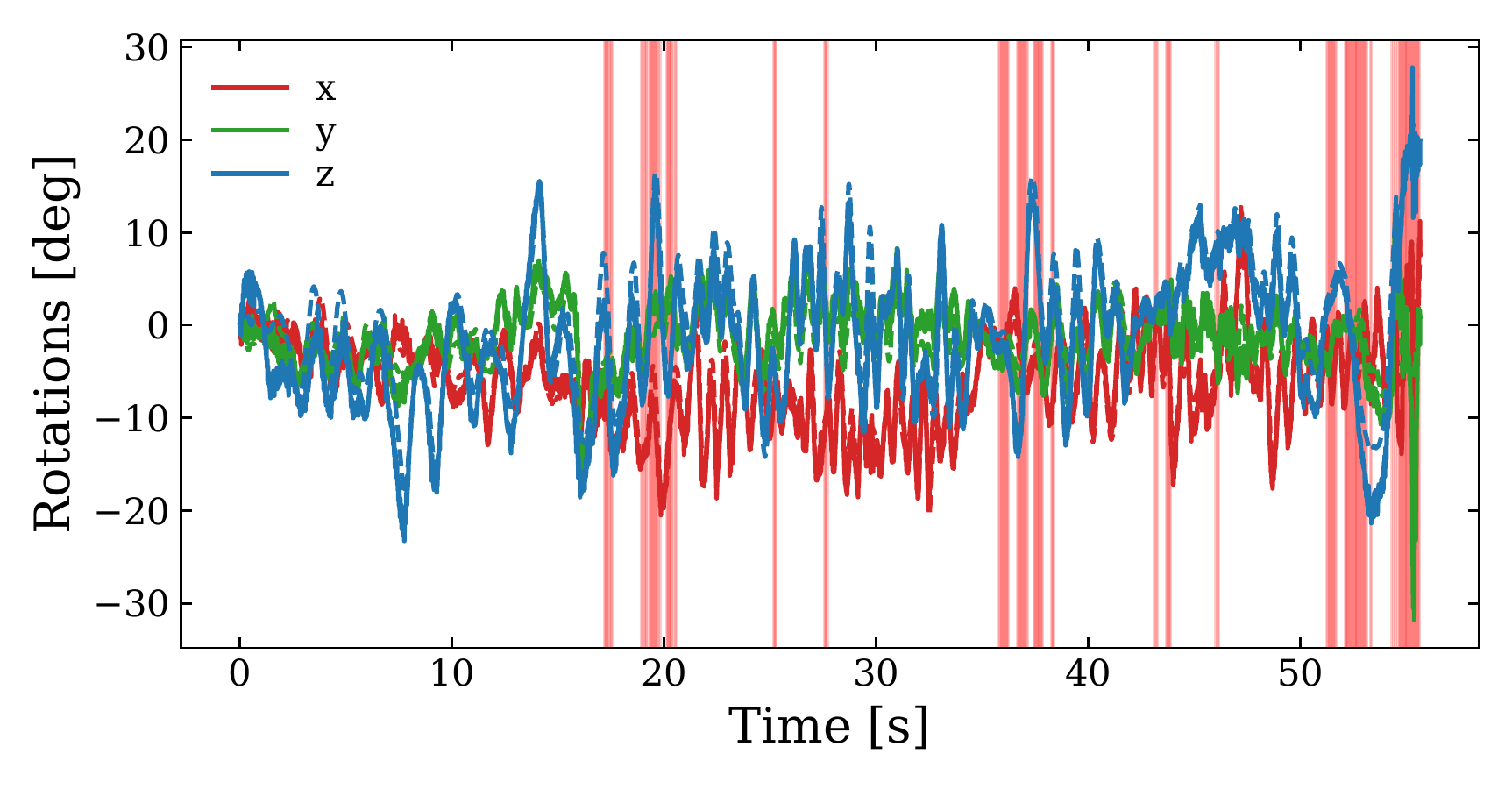}
    \caption{Estimated x, y, z rotations for ``Jumping" sequence. Solid lines show our estimate and dotted line are the ground truth. The pink region indicates that the ORB-SLAM lost track due to rapid motion.}
    \label{fig:orb-fail-orientation}
\end{figure}
\begin{figure}[!ht]
    \centering
    \includegraphics[width=\linewidth]{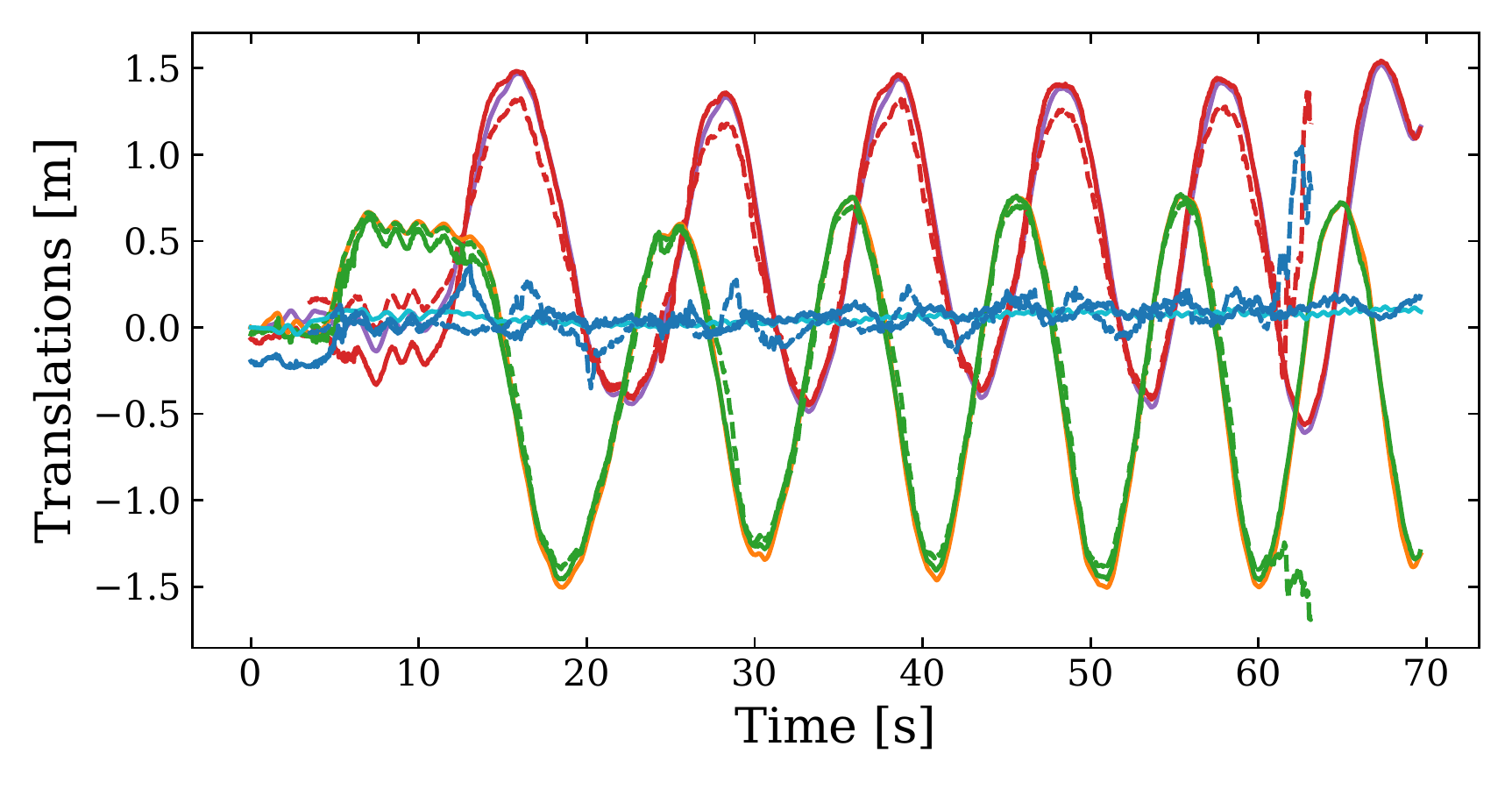}
    \caption{Estimated x, y, z translation for ``Circle" sequence. Solid lines show results from using our proposed descriptor, while dotted lines used rotated BRIEF. The estimated data x, y, z is plotted using red, green, blue and the ground truth data x, y, z is plotted using purple, orange, cyan respectively. Initialisation of rotated BRIEF version occured after our method.}
    \label{fig:brief-translation}
    \centering
    \includegraphics[width=\linewidth]{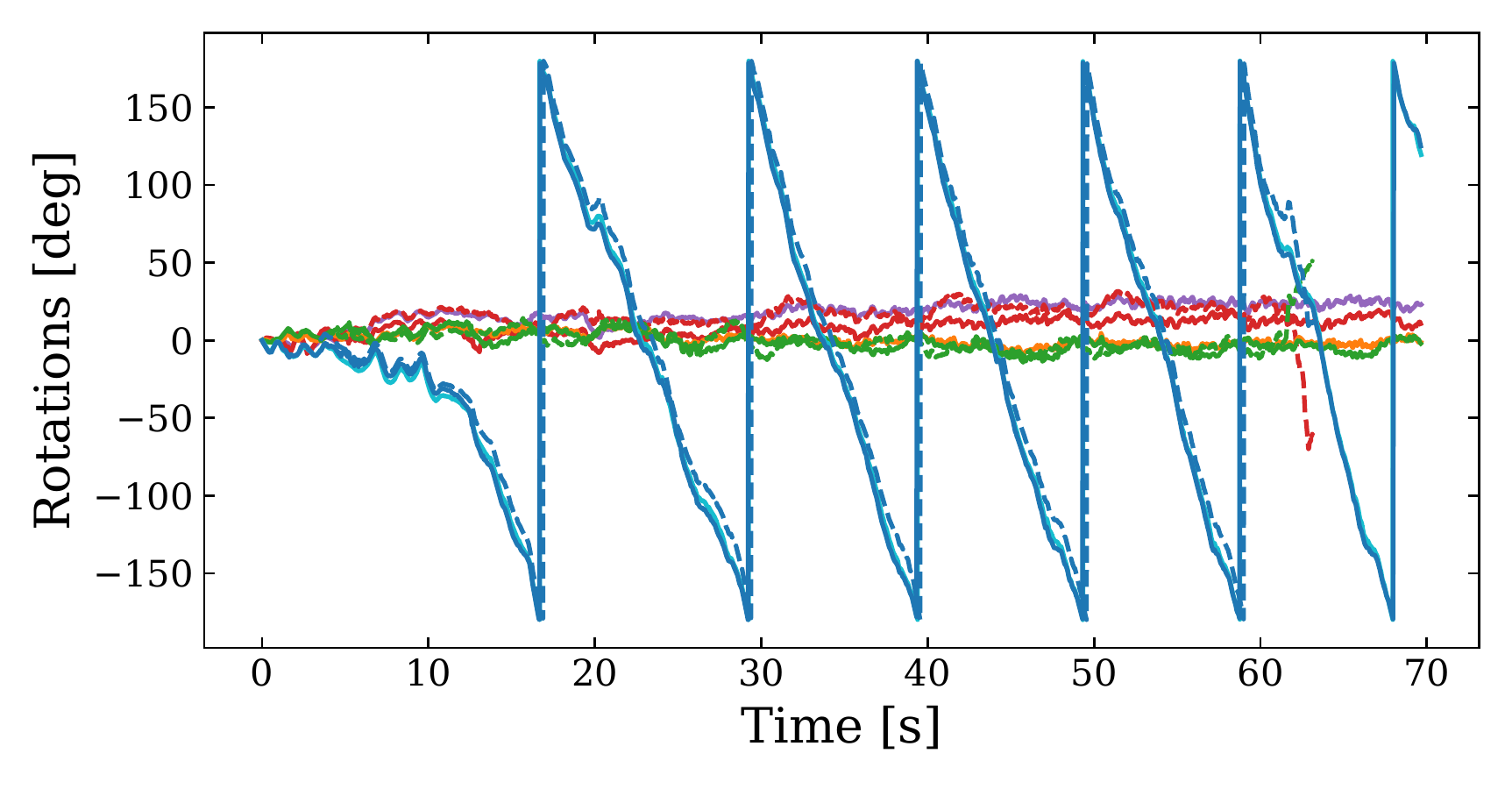}
    \caption{Estimated x, y, z rotations for ``Circle" sequence. Solid lines show results from using our proposed descriptor, while dotted lines used rotated BRIEF. 
    The estimated data x, y, z is plotted using red, green, blue and the ground truth data x, y, z is plotted using purple, orange, cyan respectively. Note rotations along z-axis wraps as full 360 degrees loops are made.}
    \label{fig:brief-orientation}
\end{figure}
\begin{figure}[!ht]
    \centering
    \includegraphics[width=\linewidth]{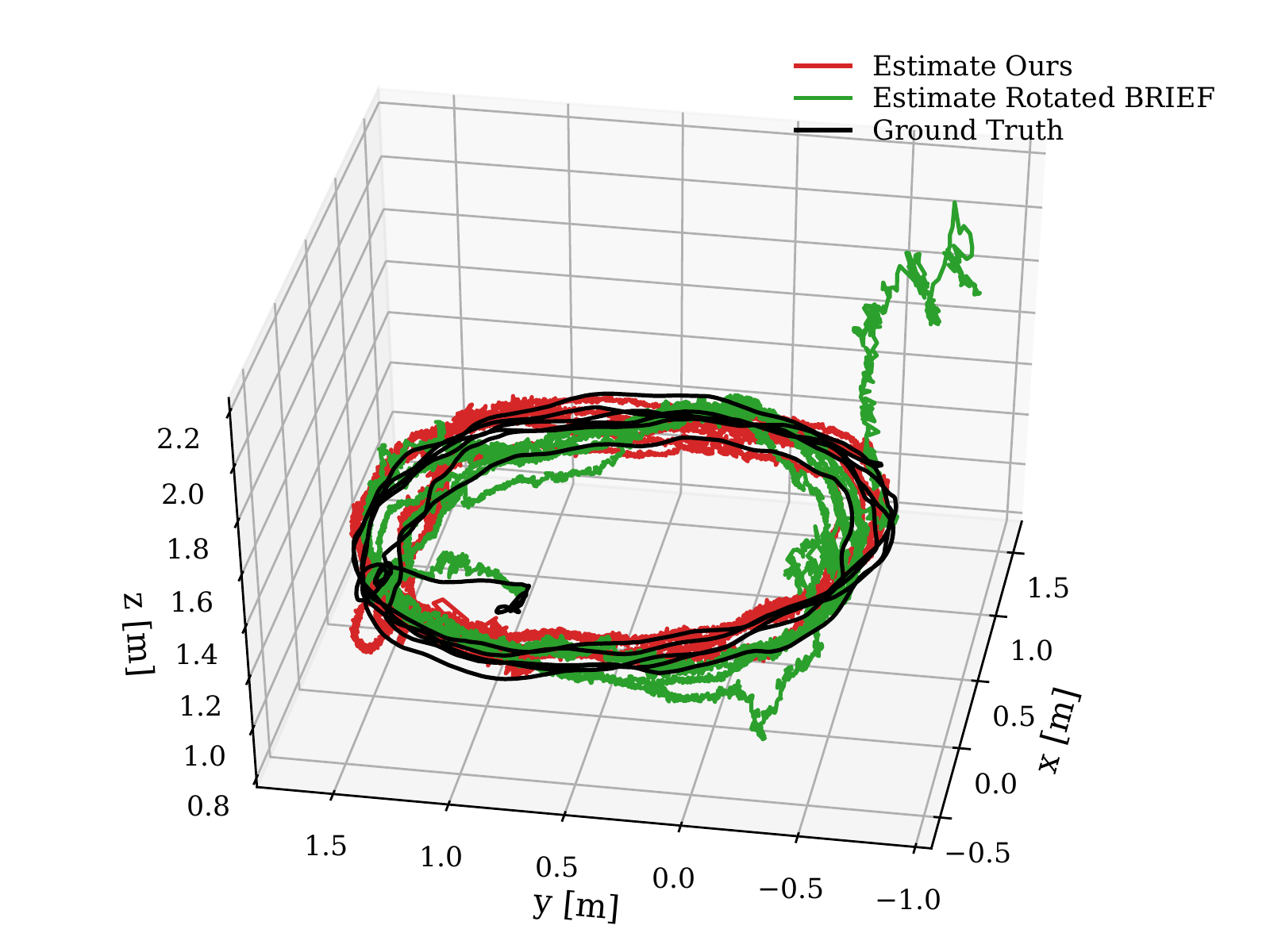}
    \caption{Estimated 3D trajectory of ``Circle" sequence using our proposed method: \Sajad{our pipeline (in red), rotated BRIEF descriptors (in green), and ground truth (in black)}.}
    \label{fig:brief-3d}
\end{figure}

\begin{figure}[!ht]
    \centering
    \includegraphics[width=\linewidth]{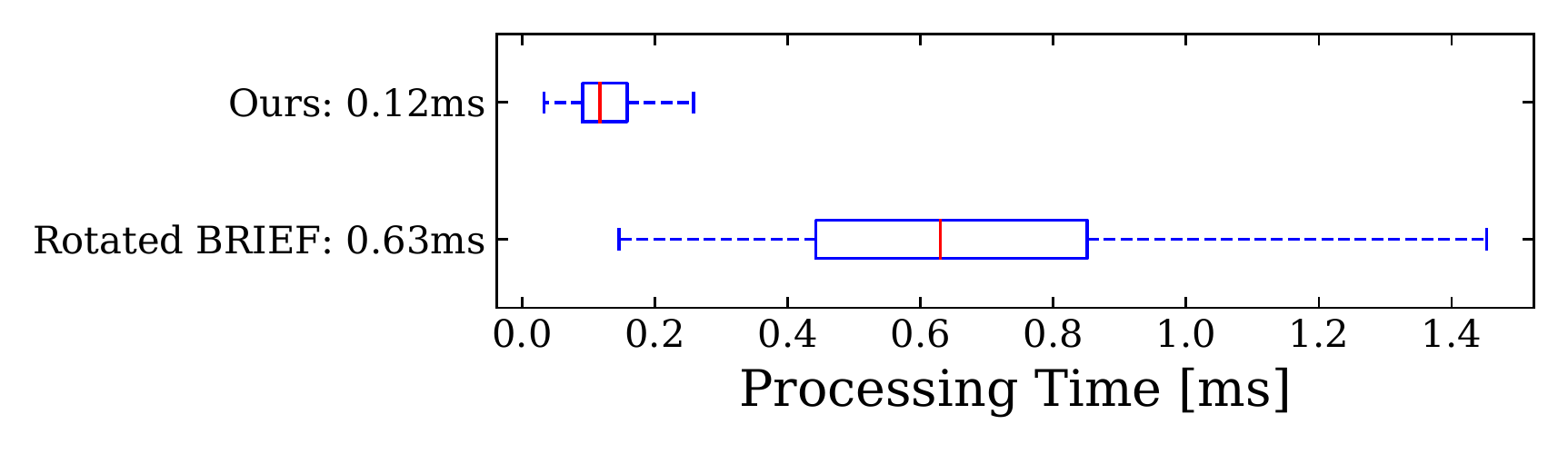}
    \caption{Comparison of the processing time of our descriptor against rotated BRIEF.}
    \label{fig:our_vs_rotated_brief}
    \centering
    \includegraphics[width=\linewidth]{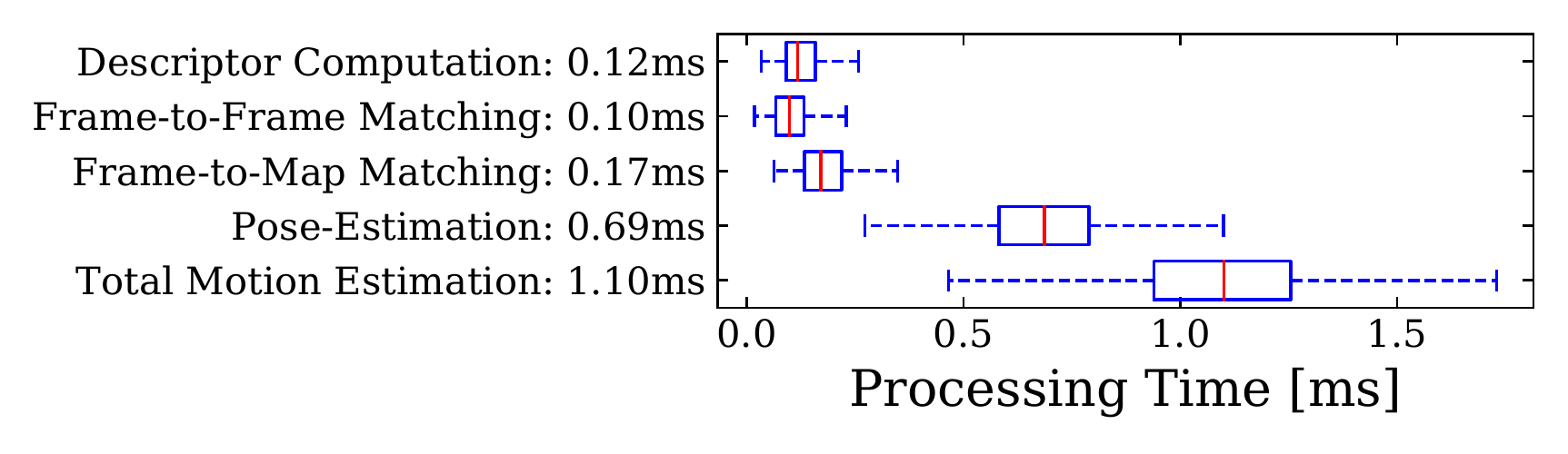}
    \caption{Breakdown of the processing time required by our motion-estimation.}
    \label{fig:motion-time}
    \centering
    \includegraphics[width=\linewidth]{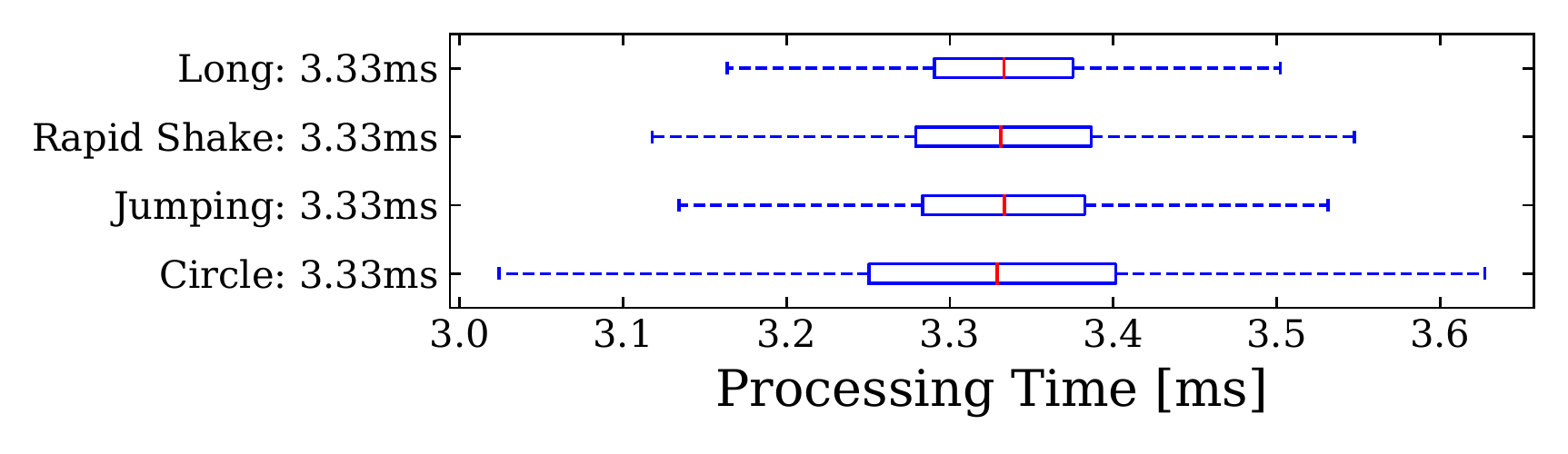}
    \caption{Processing time per frame while running the system online on different sequences. Note that the bottleneck is the SCAMP-5, which outputs features at 300 FPS.}
    \label{fig:runtime-online}
\end{figure}

\section{EXPERIMENTS}\label{sec:experiments}
We have evaluated our proposed system against ground-truth data from the Vicon motion capture system. As the our method is a monocular VO, the estimated trajectory is scaled and aligned to the ground truth data. \Sajad{Experiments have been conducted with the SCAMP-5 FPSP~\cite{dudek2005general}.} Raw intensity images are not recorded by SCAMP-5, \Sajad{because in this case, SCAMP-5 would act as a conventional camera, with a reduced frame-rate}. Thus, a direct comparison against other VO/VSLAM using a monocular camera or SCAMP-5 is not possible. \Sajad{Instead, }
a webcam was attached to SCAMP-5 to demonstrate that systems using a typical camera such as ORB-SLAM~\cite{mur-artal_orb-slam:_2015} lose track when subject to dynamic motions. 
Field of view between the two devices are different, hence, for fairness, best efforts were made to ensure both devices observe the same scene.
All \Sajad{host computations} were made on a laptop, with 4-core Intel i7-6700HQ CPU at 2.60GHz. Mapping and tracking used a single core, with visualisation, and communication with SCAMP-5 using an extra core each. 

Due to the nature of SCAMP-5, we cannot use \Paul{existing frame-by-frame video} datasets for comparison. Thus, we evaluate our system against 4 different recordings: Long, Rapid Shake, Jumping, and Circle sequences. The test scene consisted of typical tabletop objects such as desktop monitor and books. Videos of \Paul{the} live running system is available on \url{https://rmurai0610.github.io/BIT-VO}.
\begin{table}[]
\caption{Absolute Trajectory Error of different sequences, computed using evo~\cite{grupp2017evo}. The total length of the trajectory, Root Mean Square Error and Median Error is reported.}\label{table:ape}
\centering
\begin{tabular}{lcccc}
\toprule
Sequence & Length[m] &  RMSE [m] & Median [m] \\
\midrule
Long         & 68.5 & 0.108 &  0.078 \\
Rapid Shake  & 5.6  & 0.015 &  0.011 \\ 
Jumping      & 32.9 & 0.056 &  0.040 \\
Circle       & 38.3 & 0.128 &  0.084 \\
\bottomrule%
\end{tabular}
\end{table}

\begin{table}[]
\caption{Absolute Trajectory Error comparison of using our proposed descriptor and using rotated BRIEF, computed using evo~\cite{grupp2017evo}. . The total length of the trajectory, Root Mean Square Error and Median Error is reported.}\label{table:ape_brief}
\centering
\begin{tabular}{lcccc}
\toprule
Descriptor & Length[m] &  RMSE [m] & Median [m] \\
\midrule
Ours      & 38.3 & 0.128 &  0.084 \\
Rotated BRIEF & 38.3 & 0.123 &  0.107 \\\bottomrule%
\end{tabular}
\end{table}

\subsection{Accuracy and Robustness}
The ``Long'' sequence repeatedly travels the test arena for 68.5m, where many features continuously enters and leaves the sight of the SCAMP-5. Fig.~\ref{fig:long-translation} and Fig.~\ref{fig:long-orientation} illustrates the translation and rotation of our system over time. We notice a small drift along z-axis rotation; however, there are no other significant drift, with small RMSE of 0.108m as summarised in Table~\ref{table:ape} for the Absolute Trajectory Error~\cite{sturm2012benchmark}.

Similar to a 4-DoF VO for SCAMP-5~\cite{bose_visual_2017}, our system is able to track violent rotations, as shown in Fig.~\ref{fig:fast-orientation}. 
The system was subject to 4-5 shakes per second but is able to accurately track rotations along all three axes. A magnified view is provided in Fig.~\ref{fig:fast-orientation-zoom}.

\subsection{Comparison Against Visual SLAM}
In the ``Jumping" sequence, the device is subject to violent translational motions of up to 80cm caused by jumping as seen in seconds 42-48 of Fig.~\ref{fig:orb-fail-translation}.

Fig.~\ref{fig:orb-fail-translation} and Fig.~\ref{fig:orb-fail-orientation} highlights the advantage of operating at 300 FPS. Our VO pipeline is compared against ORB-SLAM~\cite{mur-artal_orb-slam:_2015} which uses data from a webcam operating at 20 FPS. The images are cropped to the resolution of SCAMP-5 which is $256\times256$.
Due to the nature of the FPSP device, frames are not recorded on the device, but rather a webcam which we have attached to the device.
The pink highlighted regions in Fig.~\ref{fig:orb-fail-translation} is where ORB-SLAM has lost track. As shown, the images captured on the webcam suffers from motion blur, while the features from SCAMP-5 does not.

\subsection{Comparison Against Other Descriptors}
An alternative option would have been to use other binary descriptors such as BRIEF~\cite{calonder2010brief} or BRISK~\cite{leutenegger2011brisk}; however, these methods use pixel intensity comparisons to build the descriptor. To use BRIEF descriptor in our setup, a small modification can be made, where the pixel intensity comparison is replaced with a XOR operation. For the rotation invariance, we follow the same approach as ORB~\cite{rublee2011orb}, with the orientation of the feature computed using the same method used by our descriptor.

To compare our descriptor against BRIEF, we have recorded the features from the SCAMP-5. The Vicon room was explored in a circular motion while pointing the camera towards the centre of the room. A modified version of rotated BRIEF from OpenCV~\cite{opencv_library} was used for the experiments, which is a 256-bit long descriptor.
Fig.~\ref{fig:brief-translation} and Fig.~\ref{fig:brief-orientation} shows that there are no major differences in the two approaches, apart from 60 seconds onward where VO using rotated BRIEF fails. Fig.~\ref{fig:brief-3d} is the 3D trajectories of our approaches together with the ground truth.
We notice that there are high-frequency noises present in our trajectories.
For each frame, the noise from analog computation causes a different set of corners to be extracted. The difference causes incorrect correspondences, thus different formulation of the optimisation problem. This results in a shaky trajectory. When the correct features are again extracted, through descriptor matching, incorrect matches are removed, thus as demonstrated our system does not build up the noise.
Table~\ref{table:ape_brief} compares the absolute trajectory error of using two different descriptors. For a fair comparison, we have excluded all measurements after 58 seconds for rotated BRIEF, such that the failed trajectory is excluded from the metric computation. We observe no significant difference in the accuracy of tracking in using either of the descriptors.
The main advantage of our descriptor is visible in Fig.~\ref{fig:our_vs_rotated_brief}. The ``Circle" sequence was executed offline using our descriptor and rotated BRIEF for 10 iterations each, and the time required to compute descriptor per frame is reported. Looking at the median, our approach is more than 5 times faster than rotated BRIEF.

\subsection{Runtime Evaluation}
Breakdown of the runtime of the motion-estimation is provided in Fig.~\ref{fig:motion-time}. The timing is measured offline over 10 iterations of the ``Circle" sequence.
Our motion estimation is highly efficient, and the median time required to estimate the pose is 1.10ms, which translates to a frame-rate of over 900 FPS.
Currently, our system does not separate map-refinement onto different thread during keyframe insertion. The median of processing time for keyframe insertion is 3.17ms, with 2.22ms, 3.98ms at 0.25, 0.75 quantile respectively. When operating at 300 FPS, time budget is only 3.33ms, thus keyframe insertion combined with motion-estimation exceeds our allowance. However, within one or two frames, the excess is resolved. For a latency critical application, it is possible to offload the keyframe insertion to a different thread. 

The runtime of the different sequences when operating the system live is reported in Fig.~\ref{fig:runtime-online}. We execute SCAMP-5 at 300 FPS, not at full capacity of 330 FPS for stable frame-rates.
As shown, our system is limited by the frame-rate of the SCAMP-5, not by our VO algorithms. ``Circle" sequence has the largest inter-quantile-range, as it required more keyframe insertions when compared to other sequences.


\section{CONCLUSION}\label{sec:conclusion}
We have presented BIT-VO, which is capable of \Sajad{performing VO} at 300 FPS \Paul{by} 
using binary edges and corners computed on the focal plane. 
Our system is simplistic and minimal, yet it is sufficient to work in challenging conditions, highlighting the advantage of operating at high effective frame rates. \Sajad{In the proposed pipeline, a robust feature matching scheme using small 44-bit descriptors was implemented. FPSP's analog computation introduces noise to the values, but the proposed method is able to distinguish the noisy features. In future, we plan to incorporate a noise model for the computation of the FPSP, to improve the accuracy of the algorithms. One of the key challenges in working with FPSPs is benchmarking of VO/VSLAM against conventional methods. If full intensity images are recorded from an FPSP for benchmarking purposes, the FPSP would not be able to operate at its high frame rate. A possible solution is to create an automated system to repeat the exact same trajectory multiple times. }

\Paul{This work will inform the design of future FPSP devices with higher computational capability, light sensitivity and pixel count.  The programmable nature of the FPSP device, in contrast to, for example, event cameras, offers the prospect of higher accuracy, and enhanced robustness through greater adaptivity.}

\section*{ACKNOWLEDGEMENTS}
This research is supported by the Engineering and Physical Sciences Research Council [grant number
EP/K008730/1]. We would like to thank Piotr Dudek, Stephen J. Carey, and Jianing Chen at the University of Manchester for kindly providing access to SCAMP-5.

\bibliographystyle{IEEEtran}
\bibliography{bib}

\end{document}